\documentclass[journal=jacsat,manuscript=article]{achemso}

\usepackage[version=3]{mhchem} 
\usepackage[utf8]{inputenc}
\usepackage{url}
\usepackage{hyperref}
\usepackage{booktabs}
\usepackage{bbding}
\usepackage{pifont}
\usepackage{wasysym}
\usepackage{utfsym}
\usepackage{fontawesome}
\usepackage{graphicx}%
\usepackage{multirow}%
\usepackage{amsmath,amssymb,amsfonts}%
\usepackage{amsthm}%
\usepackage{mathrsfs}%
\usepackage[title]{appendix}%
\usepackage{xcolor}%
\usepackage{textcomp}%
\usepackage{manyfoot}%
\usepackage{booktabs}%
\usepackage{algorithm}%
\usepackage{algorithmicx}%
\usepackage{algpseudocode}%
\usepackage{listings}%

\usepackage{xcolor}
\newcounter{bxincomm}
\definecolor{aqua}{rgb}{0.00,0.67,0.80}

\author{Yang Tan}
\altaffiliation{These authors contributed equally to this work.}
\affiliation[ECUST]{School of Information Science and Engineering, East China University of Science and Technology, 200237, China.}
\alsoaffiliation[AILAB]{Shanghai Artificial Intelligence Laboratory, 200232, China.}
\alsoaffiliation[SHCAM]{Shanghai National Center for Applied Mathematics (SJTU Center), 200240, China.}
\author{Mingchen Li}
\altaffiliation{These authors contributed equally to this work.}
\affiliation[ECUST]{School of Information Science and Engineering, East China University of Science and Technology, 200237, China.}
\alsoaffiliation[AILAB]{Shanghai Artificial Intelligence Laboratory, 200232, China.}
\alsoaffiliation[SHCAM]{Shanghai National Center for Applied Mathematics (SJTU Center), 200240, China.}

\author{Bingxin Zhou}
\altaffiliation{These authors contributed equally to this work.}
\affiliation[INS]{Institute of Natural Sciences, Shanghai Jiao Tong University, 200240, China.}
\alsoaffiliation[SHCAM]{Shanghai National Center for Applied Mathematics (SJTU Center), 200240, China.}
\author{Bozitao Zhong}
\affiliation[INS]{Institute of Natural Sciences, Shanghai Jiao Tong University, 200240, China.}
\alsoaffiliation[SHCAM]{Shanghai National Center for Applied Mathematics (SJTU Center), 200240, China.}

\author{\\Lirong Zheng}
\affiliation[INS]{Institute of Natural Sciences, Shanghai Jiao Tong University, 200240, China.}
\alsoaffiliation[UMICH]{Department of Cell and Developmental Biology \& Michigan Neuroscience Institute, University of Michigan Medical School, 48104 MI, USA.}
\author{Pan Tan}
\affiliation[INS]{Institute of Natural Sciences, Shanghai Jiao Tong University, 200240, China.}
\alsoaffiliation[AILAB]{Shanghai Artificial Intelligence Laboratory, 200232, China.}
\author{Ziyi Zhou}
\affiliation[INS]{Institute of Natural Sciences, Shanghai Jiao Tong University, 200240, China.}
\alsoaffiliation[SHCAM]{Shanghai National Center for Applied Mathematics (SJTU Center), 200240, China.}
\author{Huiqun Yu}
\affiliation[ECUST]{School of Information Science and Engineering, East China University of Science and Technology, 200237, China.}
\email{yhq@ecust.edu.cn}
\author{Guisheng Fan}
\affiliation[ECUST]{School of Information Science and Engineering, East China University of Science and Technology, 200237, China.}
\email{gsfan@ecust.edu.cn}

\author{\\Liang Hong}
\affiliation[INS]{Institute of Natural Sciences, Shanghai Jiao Tong University, 200240, China.}
\alsoaffiliation[SHCAM]{Shanghai National Center for Applied Mathematics (SJTU Center), 200240, China.}
\alsoaffiliation[ZIAS]
{Zhangjiang Institute for Advanced Study, Shanghai Jiao Tong University, 200240, China}
\alsoaffiliation[AILAB]{Shanghai Artificial Intelligence Laboratory, 200232, China.}
\email{hongl3liang@sjtu.edu.cn}

\title[SES-Adapter]
  {Simple, Efficient and Scalable Structure-aware Adapter Boosts Protein Language Models}

\abbreviations{IR,NMR,UV}

\begin{document}

\begin{tocentry}
\centering
\includegraphics[width=\linewidth]{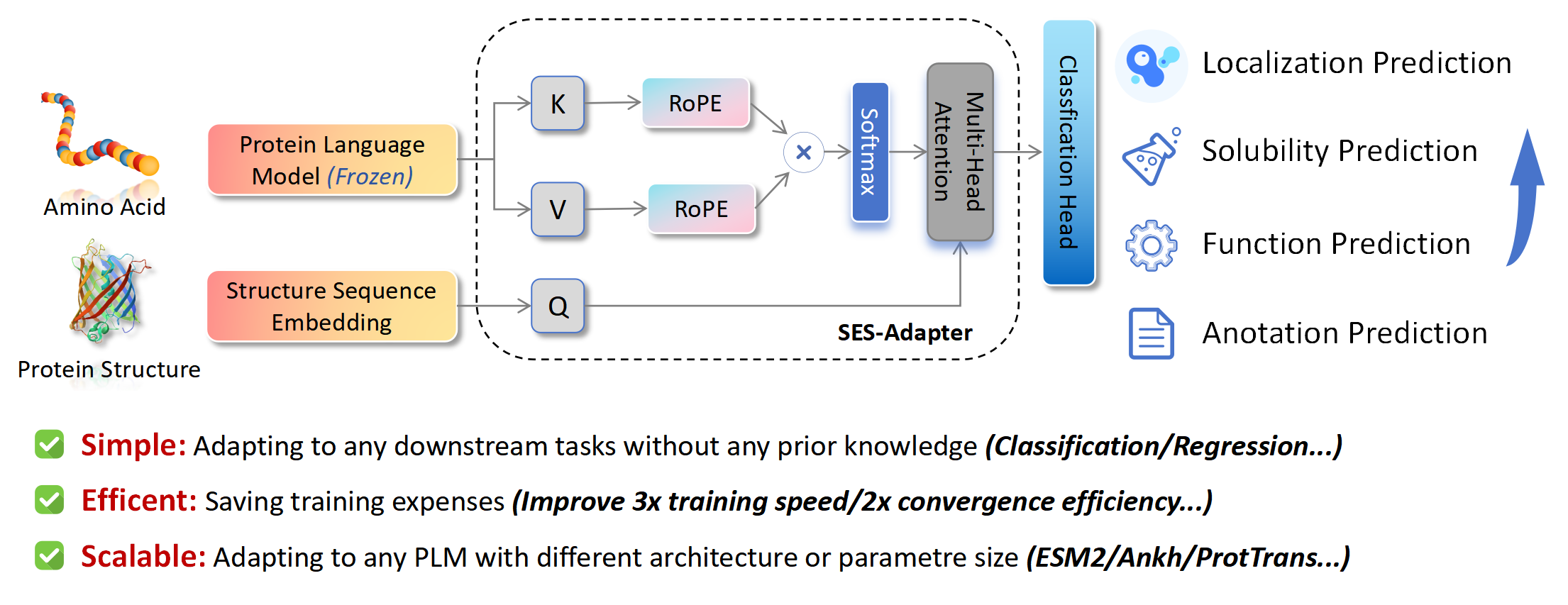}

 Workflow of SES-Adapter for PLMs to improve performance and efficiency across diverse tasks.

\end{tocentry}

\begin{abstract}
Fine-tuning Pre-trained protein language models (PLMs) has emerged as a prominent strategy for enhancing downstream prediction tasks, often outperforming traditional supervised learning approaches. As a widely applied powerful technique in natural language processing, employing Parameter-Efficient Fine-Tuning techniques could potentially enhance the performance of PLMs. However, the direct transfer to life science tasks is non-trivial due to the different training strategies and data forms. To address this gap, we introduce SES-Adapter, a simple, efficient, and scalable adapter method for enhancing the representation learning of PLMs. SES-Adapter incorporates PLM embeddings with structural sequence embeddings to create structure-aware representations. We show that the proposed method is compatible with different PLM architectures and across diverse tasks. Extensive evaluations are conducted on $2$ types of folding structures with notable quality differences, $9$ state-of-the-art baselines, and $9$ benchmark datasets across distinct downstream tasks. Results show that compared to vanilla PLMs, SES-Adapter improves downstream task performance by a maximum of $11\%$ and an average of $3\%$, with significantly accelerated training speed by a maximum of $1034\%$ and an average of $362\%$, the convergence rate is also improved by approximately $2$ times. Moreover, positive optimization is observed even with low-quality predicted structures. The source code for SES-Adapter is available at \url{https://github.com/tyang816/SES-Adapter}.
\end{abstract}

\section{Introduction}
Proteins hold significant value in biological research \cite{buccitelli2020protein_activity} and industry applications \cite{guthman2022protein_industry}. Understanding proteins and their functions has previously relied heavily on prior knowledge and extensive wet-lab experiments. Although precise, such methods are time-consuming and labor-intensive. With advancements in sequencing technology \cite{alfaro2021protein_sequencing}, there now exists a relatively abundant corpus of protein sequences, and applying model architectures from the field of natural language processing (NLP) for self-supervised training \cite{lin2023esm2, elnaggar2023ankh, elnaggar2021prottrans} become possible, thereby achieving robust representations of protein sequences. These representations enable the utilization of zero-shot or supervised downstream tasks for predicting protein properties. PLMs are fundamentally driven by the quality and quantity of data, allowing for the extraction of evolutionary commonalities and critical information from a vast array of sequences, exemplified by models such as ESM \cite{lin2023esm2, meier2021esm1v, rives2021esm1b} and ProtTrans \cite{elnaggar2021prottrans} series. They serve as end-to-end approaches to eliminate the need for specialized designs or expert knowledge, facilitating transfer learning to other prediction tasks. 

Compared to the field of NLP, the corpus of protein sequences is still limited, which in turn restricts the size of PLM parameters. Simply increasing the size of these models may not always be appropriate for the biological domain and can lead to unnecessarily high training costs \cite{elnaggar2023ankh}. Often, there is no need to train a large model from scratch for each scenario; instead, we can fine-tune pre-trained model weights with a supervised dataset specific to the target scenario, such as solubility prediction \cite{chen2023esm2_solubility, thumuluri2022netsolp}, protein-protein interaction (PPI) prediction \cite{jha2023plm_ppi1, sargsyan2024plm_ppi2}, and protein localization prediction \cite{stark2021prott5_la, thumuluri2022deeploc2}. Therefore, exploring how to enhance the representational quality of PLMs and performing efficient, lightweight fine-tuning to fully unleash their potential becomes extremely valuable. Considering the rapid updates and the numerous versions of PLMs \cite{bepler2021plm_survey}, an effective fine-tuning method also needs to be adaptable to various models.

Although protein sequence data is of high quality and abundant, the function of a protein is determined by its structure \cite{van2022protein_stc_fuc}, which contains richer and more comprehensible information. Intuitively, incorporating structural information into PLMs should improve the performance across downstream tasks. However, recent studies have indicated that crudely adding structure-aware information does not necessarily yield better results, with some research observing negative optimization in tasks like protein localization \cite{zhang2024esm-s, chen2024pst}. This might be due to high-quality training data from sequence models paired with errors in predicted structures \cite{terwilliger2024af_error}, which can decrease performance when the structure is introduced for training or prediction. Moreover, some studies suggest that introducing noise to mitigate these errors during training with predicted structures can enhance performance \cite{dauparas2022proteinmpnn, zhang2022gearnet, zhou2023protlgn}. Therefore, when utilizing protein structural information for predictions, it's important to consider the errors in predicted structures and the robustness of the model, aiming to maximize the use of key structural information while maintaining the original performance of the sequence model and filtering out noise to improve prediction outcomes. Notably, thanks to advances in folding technologies \cite{lin2023esm2, mirdita2022colabfold, jumper2021alphafold} and the availability of open-source databases \cite{burley2019rcsb, varadi2022afdb}, acquiring protein structures has become cost-effective, increasing the feasibility of integrating structural information broadly to aid fine-tuning.

In the realm of enhancing model representation through fine-tuning, there are numerous lightweight fine-tuning approaches in NLP, examples include prompt tuning \cite{lester2021prompt_tuning} and prefix tuning \cite{li2021prefix_tuning}. One mainstream approach involves updating the original model's parameters for fine-tuning, while another adds an external component to the original model for fine-tuning without altering the base model. However, most NLP fine-tuning methods are challenging to apply in the biochemistry field due to the data formation and training strategy, while the bioinformatics domain has limited and relatively rudimentary existing fine-tuning research and most methods consider amino acid sequences only. For instance, PEFT-SP \cite{zeng2023plm_peft} uses the \texttt{PEFT} library to directly fine-tune PLMs to improve signal peptide prediction. Another work simply employs LoRA \cite{hu2021lora}  for PLMs to enhance downstream task performance \cite{schmirler2023plm_lora}. Other methods include data augmentation to update original model parameters for performance improvement, such as SESNet \cite{li2023sesnet}, which uses unsupervised pseudo-labeling, and FSFP \cite{zhou2024fsfp}, which employs rank learning \cite{cao2007rank_learning} and retrieval to boost protein language model performance on zero-shot mutation prediction tasks. Some work has added graph neural network components to sequence models for downstream tasks, though these do not strictly qualify as fine-tuning methods. For example, ProtSSN \cite{tan2023protssn} initializes EGNN \cite{satorras2021egnn} with sequence models to enhance variant prediction capabilities, MIF-ST \cite{yang2023mif-st} uses CARP \cite{yang2022carp} language model to boost the inverse folding task capability of graph neural networks, and ESM-GearNet \cite{zhang2023esm-gearnet} enhances downstream task capabilities by combining with ESM2 and GearNet.

To address the challenges of the scarcity of efficient fine-tuning methods in the protein field and how to use structural information to optimize PLMs' representations without degradation, we propose SES-Adapter, a model-agnostic, structure-aware adapter that integrates language model representations with structural sequence representations through cross-modal fusion attention. For the structural sequence representations, we use FoldSeek \cite{van2023foldseek} and DSSP \cite{kabsch1983dssp} software to serialize protein structures, and convert the structural sequences into dense vectors. This adapter features a straightforward design, rapid convergence, excellent performance, error elimination, and can be extended to any model, addressing deficiencies in enhancing the transfer learning capabilities of protein language models and surpassing most specialized methods designed for specific downstream tasks. To validate the SES-Adapter's versatility, it was adapted to nine state-of-the-art baselines across the ESM2 \cite{lin2023esm2}, ProtBert \cite{elnaggar2021prottrans}, ProtT5 \cite{elnaggar2021prottrans}, and Ankh \cite{elnaggar2023ankh} series, and extensively evaluation on nine datasets for tasks including protein localization prediction, solubility prediction, function prediction, and annotation prediction. The experiments demonstrated that the SES-Adapter outperformed vanilla PLMs, with a maximum performance increase of $11\%$ and an average of $3\%$; training speeds were enhanced by up to $1034\%$ and an average of $362\%$, with an approximate $2$ times improvement in convergence efficiency. Additionally, to confirm that the SES-Adapter's serialization strategy effectively mitigates potential prediction errors and is insensitive to structural quality, we conducted comparative tests using structures folded by ESMFold \cite{lin2023esm2} and AlphaFold2 \cite{jumper2021alphafold}. The results showed the performance difference between the two types of structures is up to $0.6\%$, verifying that this method can effectively overcome structural inaccuracies and avoid the negative optimization issues associated with using predicted or low-quality structures.

\section{Materials and Methodology}
\subsection{Dataset}

\begin{table}[t]
\caption{Summary of Benchmarks for downstream tasks. We report \textit{mean (standard deviation)} pLDDT scores of folded proteins for each dataset.} 
\label{tab:dataset}
\resizebox{\textwidth}{!}{
    \begin{tabular}{@{}ccccccc@{}}
    \toprule
    Dataset & AF2$\_$pLDDT & EF$\_$pLDDT & \# Train & \# Valid & \# Test & Metrics \\ \midrule
    \multicolumn{7}{c}{\textbf{Localization Prediction}} \\ \midrule
    DeepLocBinary (DLB) & $79.57_{(12.06)}$ & $77.10_{(14.62)}$ & $5,735$ & $1,009$ & $1,728$ & ACC \\
    DeepLocMulti (DLM) & $77.34_{(12.77)}$ & $74.88_{(15.23)}$ & $9,324$ & $1,658$ & $2,742$ & ACC \\ \midrule
    \multicolumn{7}{c}{\textbf{Solubility Prediction}} \\ \midrule
    DeepSol (DS) & - & $79.59_{(13.36)}$ & $62,478$ & $6,942$ & $2,001$ & ACC \\ \midrule
    DeepSoluE (DSE) & - & $80.68_{(12.79)}$ & $10,290$ & $1,143$ & $3,100$ & ACC \\ \midrule
    \multicolumn{7}{c}{\textbf{Function Prediction}} \\ \midrule
    MetalIonBinding (MIB) & $92.36_{(6.43)}$ & $83.66_{(8.73)}$ & $5,068$ & $662$ & $665$ & ACC \\
    Thermostability (Thermo) & $79.02_{(12.26)}$ & $74.60_{(13.82)}$ & $5,056$ & $639$ & $1336$ & Spearman’s $\rho$ \\ \midrule
    \multicolumn{7}{c}{\textbf{Annoation Prediction}} \\ \midrule
    GO-MF (MF) & $91.77_{(6.68)}$ & $82.84_{(9.68)}$ & $22,081$ & $2,432$ & $3,350$ & Fmax \\
    GO-BP (BP) & $91.35_{(7.06)}$ & $82.00_{(10.65)}$ & $20,947$ & $2,334$ & $3,350$ & Fmax \\
    GO-CC (CC) & $90.07_{(8.05)}$ & $79.57_{(11.61)}$ & $9,552$ & $1,092$ & $3,350$ & Fmax \\ \bottomrule
    \end{tabular}
}
\end{table}

Our benchmark comprises 9 datasets across 4 tasks, with all proteins folded using ESMFold to obtain their structures. Except for the Solubility prediction task, datasets for other tasks also include structures obtained from AlphaFold2 database \cite{varadi2022afdb}, as detailed in Table \ref{tab:dataset}.

\subsubsection{Protein Localization Prediction} 
Predicting the specific intracellular location of proteins can unveil their biological functions, inform disease treatments, and facilitate drug development \cite{real2009Alzheimer}. We utilized both a multi-class and a binary-class dataset from DeepLoc \cite{almagro2017deeploc}, with a deduplication process that removed 30\% of sequence similarities. DeepLocBinary (DLB) aims to ascertain whether a protein is a membrane-bound protein, for which we divided a new training/validation at a random 4:1 ratio within the training dataset. The DeepLocMulti (DLM) dataset encompasses 10 potential locations; proteins situated in the lamina, chromosome, or nucleus speckle, for example, are predicted to be in the ``Nucleus". The division of the dataset is taken from LAProtT5 \cite{stark2021prott5_la}.

\subsubsection{Protein Solubility Prediction}
Protein solubility is a prerequisite for expression, while solubility defects can lead to protein aggregation, thereby affecting protein bioactivity, hindering protein-based drug development, and causing a variety of diseases \cite{chiti2017sol_disease}. We used two binary-class datasets, derived from DeepSol (DS) \cite{khurana2018deepsol} and DeepSoluE (DSE) \cite{wang2023deepsolue}, respectively. The division of the training/validation sets and test sets remains unchanged, except for the exclusion of proteins that could not be folded using ESMFold.

\subsubsection{Protein Function Prediction} 
The functionality of proteins encompasses numerous aspects. To streamline our study, we employed two datasets: MetalIonBinding (MIB) and Thermostability (Thermo), which are focused on binary classification and regression fitness predictions, respectively. The MetalIonBinding task aims to predict the existence of metal ion-binding sites in proteins, with the dataset sourced from Revisiting-PLMs \cite{hu2022revisiting-plms}. Additionally, for the Thermostability task, we utilize the ``Human-cell" split from FLIP \cite{dallago2021flip}, implementing min-max normalization on the labels.

\subsubsection{Protein Annotation Prediction}
The transfer learning of protein function annotation can assist in identifying the functions of unknown proteins, reducing the cost of experimental trial and error and enhancing efficiency \cite{szklarczyk2023protein_annotation}. Our dataset is derived from Gene Ontology (GO) terms prediction, where the GO benchmark comprises three branches: Molecular Function (MF), Biological Process (BP), and Cellular Component (CC). The dataset and its divisions are sourced from DeepFRI \cite{gligorijevic2021deepfri}, used for predicting multi-label multi-class protein annotation information.

\begin{figure}[!t]
    \centering
    \includegraphics[width=\textwidth]{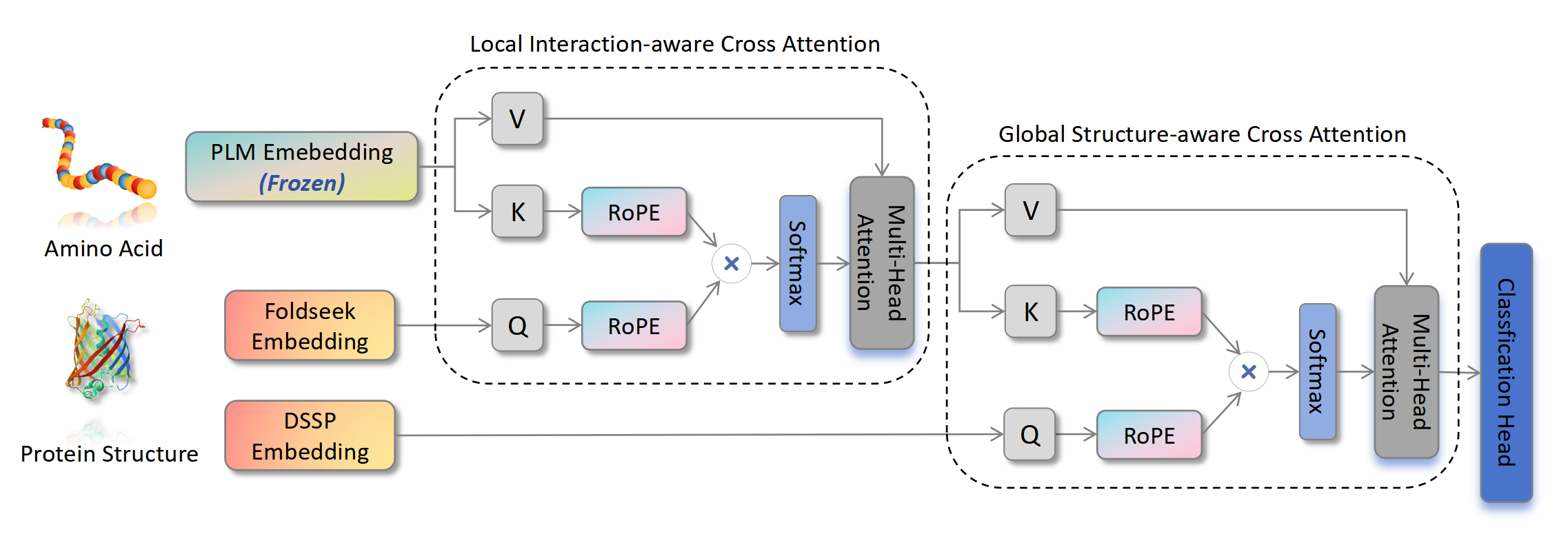}
    \caption{Architecture of SES-Adapter. Amino acid sequences are inputted into a PLM and protein structures are serialized using FoldSeek and DSSP, then projected to obtain embeddings. When using the full SES-Adapter, initial cross-modal attention is performed with the FoldSeek sequence, and the representation obtained is then subjected to another cross-modal attention with the DSSP sequence. The final embeddings are fed into a classification head for predicting downstream tasks. These two components are optional; when both are omitted, the setup defaults to a vanilla PLM. }
    \label{fig:framework}
\end{figure}

\subsection{SES-Adapter Architecture}

SES-Adapter architecture is depicted in Figure \ref{fig:framework} and consists of three main steps, encoding stage, feature fusion, and label prediction. During the encoding stage, a PLM encodes amino acid sequences into semantic vectors. The protein structure is then processed using FoldSeek and DSSP to generate local structural sequences and secondary structural sequences, respectively, which are subsequently projected to create two structure-aware vectors. In the feature fusion stage, embeddings from the PLMs are combined with these two types of structural embeddings using a cross-modal multi-head attention mechanism with rotary positional encoding (RoPE) \cite{su2024roformer} to integrate both local interaction and overall structural information. It's important to note that the inclusion of these two structural components is optional, and one may choose to utilize only one type of structural sequence. Finally, in the prediction stage, the structure-informed vectors are pooled and fed into a classification head for downstream task prediction. 

\subsection{Embedding Generation}

For a given protein sequence, input into a PLM yields an output from the last layer that serves as a representation vector of dimensions $l\times d$, where $l$ represents the sequence length and $d$ is the hidden variable dimension number of the PLM. For instance, processing a sequence of $200$ amino acids with \textbf{ESM2-650M} results in a $200\times 1280$ vector. The PLM is frozen during downstream task training, meaning its gradients do not update. For protein structures, 20-dimensional 3Di token sequences and 8-dimensional secondary structure sequences are generated using FoldSeek and DSSP, respectively. Since the alphabets for structural sequences overlap with those of amino acid sequences, we directly use the amino acid vocabulary to tokenize these sequences into one-hot vectors, which are then fed into an embedding layer to produce local interaction and global structure embeddings. These embeddings are shaped $l\times d$ maintaining consistency with the PLM's output dimensions. 

\subsection{Multi-Head Cross Modality Attention}

In the feature fusion stage, we employed a refined cross-modal multi-head attention method, drawing inspiration from ESM2's approach of using RoPE in place of traditional relative position encoding. During cross-modal interaction, PLM embeddings serve as both key and value, while structural embeddings function as the query. Since the key and value representations originate from the same source, we simply pass the query and key through RoPE before calculating the attention scores.

The amino acid or structural token embedding $\mathbb{S}_N$ can be denoted as $\mathbb{E}_N=\{\boldsymbol{x}_i\}^N_{i=1}$, where $\boldsymbol{x}_i\in\mathbb{R}^d$ is a d-dimensional word embedding vector of token $\boldsymbol{w}_i$ at position $i$ but without positional information. $\boldsymbol{x}_m$ is structural embedding vector and $\boldsymbol{x}_n$ is PLM embedding vector, and $m$ and $n$ are their absolute positions. 
\begin{align}
\boldsymbol{Q}_m=f_q(\boldsymbol{x}_m,m),\quad \boldsymbol{K}_n=f_k(\boldsymbol{x}_n,n),
\end{align}
The function $f$ projects the word embedding and its position into $\boldsymbol{Q}_m$ and $\boldsymbol{K}_n$, respectively.

The RoPE function applies a rotational transformation to each pair of dimensions from the input vectors $\boldsymbol{x}_m$ and $\boldsymbol{x}_n$ based on a rotation angle $\theta$ that is dependent on the relative positional difference $m - n$.
\begin{equation}
\langle f_q(\boldsymbol{x}_m,m),f_k(\boldsymbol{x}_n,n)\rangle=\text{RoPE}(\boldsymbol{x}_m,\boldsymbol{x}_n,m-n).
\end{equation}
\begin{equation}
    \begin{aligned}
    \mathrm{RoPE}(\boldsymbol{x}_m,\boldsymbol{x}_n,m-n)=&\sum_{i=0}^{d/2-1}(\cos(\theta)\boldsymbol{x}_{2i,m}\boldsymbol{x}_{2i,n}+\sin(\theta)\boldsymbol{x}_{2i,m}\boldsymbol{x}_{2i+1,n}\\&-\sin(\theta)\boldsymbol{x}_{2i+1,m}\boldsymbol{x}_{2i,n}+\cos(\theta)\boldsymbol{x}_{2i+1,m}\boldsymbol{x}_{2i+1,n})
    \end{aligned}
\end{equation}
where $\theta=\frac{m-n}{10000^{2i/d}}$ is the rotation angle computed based on the relative position $m - n$, dimension index $i$ and dimension number $d$. $\boldsymbol{x}_{2i,\{m, n\}}$ and $\boldsymbol{x}_{2i+1,\{m, n\}}$ are the components of vector $\boldsymbol{x}_{\{m, n\}}$ at dimensions $2i$ and $2i+1$, respectively. 

Then the whole sequence query, key representation termed $\boldsymbol{Q}$, $\boldsymbol{K}$ with positional information, and value representation $\boldsymbol{V}$ will be fed into the scale attention machine.
\begin{align}
\text{Attention}(\boldsymbol{Q},\boldsymbol{K},\boldsymbol{V})=\text{softmax}\left(\frac{\boldsymbol{Q}\boldsymbol{K}^\intercal}{\sqrt{d}}\right)\boldsymbol{V}
\label{eq:2}
\end{align}

Partitioning the attention mechanism into multiple heads, thereby generating several subspaces, enables the model to attend to diverse facets of the information. 
\begin{equation}
\text{MultiHead}(\boldsymbol{Q},\boldsymbol{K},\boldsymbol{V})=\text{Concat}({\rm head_1},...,{\rm head_n})\boldsymbol{W}^O \label{eq3}
\end{equation}
where ${\rm head_i}=\text{Attention}(\boldsymbol{Q}_i,\boldsymbol{K}_i,\boldsymbol{V}_i)$ and $i$ indicates the $i_{th}$ attention head, $\boldsymbol{W}^O$ is a linear projection matrix. The final mapped structure-aware representations can then be used for predicting downstream tasks.

\subsection{Classification Head and Training Objective}

It's worth noting that we use the classical mean pooling classification head in our comparisons. Two layers of linear projection with a dropout layer and GeLU activation function are used. For the classification tasks, we utilize Cross-Entropy as the loss function. For regression tasks, such as Thermostability, we employ Mean-Squared-Error.

\subsection{Experimental Setups}

\subsubsection{Training Setups}

We used the AdamW \cite{loshchilov2017adam} optimizer with a learning rate set at $0.0005$, a weight decay of $0.01$, and a dropout rate of $0.1$ for the output layer. To ensure stable training costs and avoid issues such as memory explosion, we adopted a dynamic batching approach, filling each batch up to $60,000$ tokens to ensure $n \times l \leq 60,000$, where $n$ is the number of sequences and $l$ is the max length of sequences at the current batch. Due to differences in data volume and training efficiency, we applied different early stop settings for different tasks. For the \textit{Solubility Prediction} task, the maximum training epoch was set to $10$ with a patience of $3$. The \textit{Annotation Prediction} task had a maximum training epoch of $50$ with a patience of $5$. For the remaining tasks, the maximum epoch was set to $15$ with a patience of $5$. The monitor for early stopping for all tasks is based on validation metrics, such as accuracy (ACC) for DeepLocBinary. All experiments and protein folding with ESMFold were conducted on eight 80GB-VRAM A800 GPUs.

\subsubsection{Evaluation Metrics}
Some evaluation metrics are reported in our experiments to evaluate the performance of different models, including accuracy (ACC), maximum F1-Score (Fmax), Matthew's correlation coefficient (MCC), and Spearman’s $\rho$, all the computation metrics are derived from the TorchMetrics library \cite{detlefsen2022torchmetrics}, except the Fmax is from \texttt{TorchDrug} \cite{zhu2022torchdrug}. Their calculation equations are as follows:
\begin{equation}
    \mathrm{ACC}=\frac{\mathrm{TN}+\mathrm{TP}}{\mathrm{TN}+\mathrm{TP}+\mathrm{FN}+\mathrm{FP}}
\end{equation}
\begin{equation}
    \mathrm{F1}=\frac{2\times\mathrm{TP}}{2\times\mathrm{TP}+\mathrm{FP}+\mathrm{FN}}
\end{equation}
\begin{equation}
    \begin{aligned}\mathrm{MCC}&=\frac{\mathrm{TP}\cdot\mathrm{TN}-\mathrm{FP}\cdot\mathrm{FN}}{\sqrt{(\mathrm{TP}+\mathrm{FN})\cdot(\mathrm{TP}+\mathrm{FP})\cdot(\mathrm{TN}+\mathrm{FN})\cdot(\mathrm{TN}+\mathrm{FP})}}\end{aligned}
\end{equation}
\begin{equation}
    \rho=1-\frac{6\sum d_i^2}{n(n^2-1)}
\end{equation}
where TP, TN, FP, and FN represent the numbers of true positives, true negatives, false positives, and false negatives, respectively. The Fmax metric is a measure of model performance that seeks the optimal threshold to maximize F1 scores.

\section{Results and Analysis}

\subsection{SES-Adapter Improves Performance Across Diverse Tasks}

\begin{table}[]
\centering
\resizebox{0.6\textwidth}{!}{%
\begin{tabular}{@{}ccccc@{}}
\toprule
Type & Model & Version & Params & Embed. Dim \\ \midrule
\multirow{5}{*}{Encoder-Only} & \multirow{3}{*}{ESM2} & t30 & 150M & 640 \\
 &  & t33 & 652M & 1,280 \\
 &  & t36 & 3,000M & 2,560 \\ \cmidrule(l){2-5} 
 & \multirow{2}{*}{ProtBert} & uniref & 420M & 1,024 \\
 &  & bfd & 420M & 1,024 \\ \midrule
\multirow{4}{*}{Encoder-Decoder} & \multirow{2}{*}{ProtT5} & xl\_uniref & 3,000M & 1,024 \\
 &  & xl\_bfd & 3,000M & 1,024 \\ \cmidrule(l){2-5} 
 & \multirow{2}{*}{Ankh} & base & 450M & 768 \\
 &  & large & 1,150M & 1,536 \\ \bottomrule
\end{tabular}%
}
\caption{PLM description.}
\label{tab:model}
\end{table}

\begin{table}[]
\caption{
Average performance under different settings of SES-Adapter across nine PLMs and nine datasets. ``SES." indicates the usage of either SES-Adapter or the vanilla PLM. 
}
\label{tab:main}
\resizebox{\textwidth}{!}{
\begin{tabular}{@{}ccccccccclll@{}}
\toprule
\multirow{3}{*}{Model} & \multirow{3}{*}{Version} & \multirow{3}{*}{SES.} & \multicolumn{2}{c}{Localization} & \multicolumn{2}{c}{Solubility} & \multicolumn{2}{c}{Function} & \multicolumn{3}{c}{Annotation} \\ \cmidrule(l){4-12} 
 &  &  & DLB & DLM & DS & DSE & MIB & Thermo & \multicolumn{1}{c}{MF} & \multicolumn{1}{c}{BP} & \multicolumn{1}{c}{CC} \\ \cmidrule(l){4-12} 
 &  &  & ACC & ACC & ACC & ACC & ACC & Sp. $\rho$ & \multicolumn{1}{c}{Fmax} & \multicolumn{1}{c}{Fmax} & \multicolumn{1}{c}{Fmax} \\ \midrule
\multirow{6}{*}{ESM2} & \multirow{2}{*}{t30} & \ding{55} & 91.2 & 77.02 & 63.37 & 55.12 & 68.57 & 67.69 &  56.44&  41.21&  48.11\\
 &  & \checkmark & \textbf{93.23} & \textbf{79.52} & \textbf{74.29} & \textbf{55.20} & \textbf{72.23} & \textbf{69.24} &  \textbf{61.68}&  \textbf{44.25}&  \textbf{50.88}\\ \cmidrule(l){2-12} 
 & \multirow{2}{*}{t33} & \ding{55} & 91.84 & 81.33 & \multicolumn{1}{l}{63.17} & 55.97 & 67.97 & 67.31 &  60.80&  45.46&  51.23\\
 &  & \checkmark & \textbf{93.32} & \textbf{82.35} & \multicolumn{1}{l}{\textbf{71.57}} & \textbf{56.15} & \textbf{71.33} & \textbf{69.47} &  \textbf{62.32}&  \textbf{46.26}&  \textbf{52.75}\\ \cmidrule(l){2-12} 
 & \multirow{2}{*}{t36} & \ding{55} & 90.57 & 80.82 & 64.17 & 54.58 & 70.24 & 68.53 & \multicolumn{1}{c}{-} & \multicolumn{1}{c}{-} & \multicolumn{1}{c}{-} \\
 &  & \checkmark & \textbf{93.33} & \textbf{82.57} & \textbf{70.88} & \textbf{54.75} & \textbf{71.61} & \textbf{68.96} & \multicolumn{1}{c}{-} & \multicolumn{1}{c}{-} & \multicolumn{1}{c}{-} \\ \midrule
\multirow{4}{*}{ProtBert} & \multirow{2}{*}{uniref} & \ding{55} & 87.09 & 74.14 & 63.02 & 54.24 & 64.96 & 65.35 &  46.75&  38.24&  49.89\\
 &  & \checkmark & \textbf{91.50} & \textbf{75.17} & \textbf{72.28} & \textbf{54.65} & \textbf{68.45} & \textbf{65.60} &  \textbf{52.79}&  \textbf{39.16}&  \textbf{51.31}\\ \cmidrule(l){2-12} 
 & \multirow{2}{*}{bfd} & \ding{55} & 89.01 & 75.2 & 64.32 & 54.97 & 65.41 & 65.28 &  46.59&  38.94&  50.05\\
 &  & \checkmark & \textbf{92.11} & \textbf{77.23} & \textbf{73.69} & \textbf{55.90} & \textbf{69.75} & \textbf{66.14} &  \textbf{57.67}&  \textbf{41.33}&  \textbf{51.15}\\ \midrule
\multirow{4}{*}{ProtT5} & \multirow{2}{*}{xl\_uniref}& \ding{55} & 92.25 & 82.02 & 67.07 & 55.03 & 75.24 & 68.36 & \multicolumn{1}{c}{-} & \multicolumn{1}{c}{-} & \multicolumn{1}{c}{-} \\
 &  & \checkmark & \textbf{93.31} & \textbf{83.66} & \textbf{73.84} & \textbf{55.43} & \textbf{75.97} & \textbf{69.08}& \multicolumn{1}{c}{-} & \multicolumn{1}{c}{-} & \multicolumn{1}{c}{-} \\ \cmidrule(l){2-12} 
 & \multirow{2}{*}{xl\_bfd} & \ding{55} & 91.72 & 78.88 & 66.17 & 55.32 & 73.14 & 67.29 & \multicolumn{1}{c}{-} & \multicolumn{1}{c}{-} & \multicolumn{1}{c}{-} \\
 &  & \checkmark & \textbf{92.72} & \textbf{79.93} & \textbf{72.98} & \textbf{55.57} & \textbf{73.86} & \textbf{67.99} & \multicolumn{1}{c}{-} & \multicolumn{1}{c}{-} & \multicolumn{1}{c}{-} \\ \midrule
\multirow{4}{*}{Ankh} & \multirow{2}{*}{base} & \ding{55} & 89.76 & 78.48 & 62.57 & 55.15 & 72.64 & 68.33 &  57.34&  42.23&  50.85\\
 &  & \checkmark & \textbf{93.33} & \textbf{81.01} & \textbf{72.53} & \textbf{55.36} & \textbf{73.00} & \textbf{69.47} &  \textbf{65.06}&  \textbf{47.47}&  \textbf{52.20}\\ \cmidrule(l){2-12} 
 & \multirow{2}{*}{large} & \ding{55} & 89.93 & 78.81 & 63.72 & 54.25 & 73.84 & 67.52 &  55.64&  41.72&  52.77\\
 &  & \checkmark & \textbf{93.32} & \textbf{83.40} & \textbf{72.84} & \textbf{54.94} & \textbf{76.57} & \textbf{68.87} &  \textbf{64.61}&  \textbf{46.96}&  \textbf{53.68}\\ \bottomrule
\end{tabular}
}
\end{table}

We evaluate the efficacy of incorporating structural information using nine different language models with varying parameter sizes and training data across nine different datasets. The baseline models include ESM2, ProtBert, ProtT5, and Ankh series. The first two types of models employ an encoder-only architecture, while the latter two feature a complete encoder-decoder structure, using T5 architecture \cite{raffel2020t5} and an asymmetric Transformer, respectively. However, we only utilized the outputs from their encoders as PLM embeddings. Details on the specific PLMs used can be found in Table \ref{tab:model}, the PLM embedding dimension determines the trainable parameters of the SES-Adapter. The implementation of PLMs is shown in Table \textcolor{blue}{S1}. Given that datasets for annotation prediction converge slowly and incur high training costs, we did not conduct tests on baseline models with 3 billion parameters for MF, BP, and CC.

In Table \ref{tab:main}, the scores for SES-Adapter are derived from using datasets from AlphaFold2 and ESMFold, as well as from scenarios using FoldSeek and DSSP separately, and simultaneously using both structural information, averaging six scores. However, for the Solubility dataset, only ESMFold structures are available, so the average is based on three scores. Table \ref{tab:main} shows that after incorporating structure-based fine-tuning, performance improved across each dataset compared to using just sequence information. Interestingly, a notable bifurcation was observed in solubility prediction; the DeepSol dataset showed significant improvements, in some cases up to $10\%$, while the gains on the DeepSoluE dataset were minimal, about $1\%$, possibly due to the size difference—DeepSol's training set is six times larger than DeepSoluE's. For the Localization prediction task, improvements ranged between $1\%$ and $3\%$; for the function prediction task, the gains were more pronounced, between $2\%$ and $5\%$. As for the annotation prediction task, the improvements for GO-CC were between $1\%$ and $3\%$, while GO-MF and GO-BP exhibited wider fluctuations, ranging from $1\%$ to $11\%$ and $1\%$ to $5\%$ respectively. The main reason for this variation is the choice of PLM, with some models that have weaker representational capabilities for these tasks showing more significant improvements when supplemented with SES-Adapter.

In addition, we also conduct a comparison of three fine-tuning methods: the first involves freezing the language model and fine-tuning the classification head using only the sequence; the second adjusts all parameters of both the language model and the classification head using sequence information. The first two methods are thus named Seq-Tuning because they solely utilize sequence information. The third method, which we propose, is structural fine-tuning. It freezes the language model and fine-tunes the cross-modal attention head using structural sequences. As can be seen in Table \ref{tab:fine-tuning_strategy}, under the same PLM, dataset, and hyperparameter settings, the SES-Adapter, with only a modest increase in training parameters, significantly outperforms the comprehensive fine-tuning methods.

\begin{table}[t]
\caption{Experimental comparison with two types of fine-tuning strategy on ProtBert and ESM-1b.}
\label{tab:fine-tuning_strategy}
\resizebox{0.9\textwidth}{!}{
\begin{tabular}{@{}cccccc@{}}
\toprule
Model                     & Params      & Strategy    & DeepLocBinary  & DeepLocMulti   & DeepSol        \\ \midrule
\multirow{3}{*}{ESM-1b}   & 1.64M/652M  & Seq-Tuning  & 91.61          & 79.94          & 67.02          \\
                          & 652M/652M   & Seq-Tuning  & 92.4           & 78.13          & 70.23          \\
                          & 14.82M/652M & SES-Adapter & \textbf{92.48} & \textbf{82.64} & \textbf{71.21} \\ \midrule
\multirow{3}{*}{ProtBert} & 1.05M/420M  & Seq-Tuning  & 87.09          & 74.14          & 59.17          \\
                          & 420M/420M   & Seq-Tuning  & 91.32          & 76.53          & 68.15          \\
                          & 9.50M/420M  & SES-Adapter & \textbf{92.42} & \textbf{77.72} & \textbf{72.96} \\ \bottomrule
\end{tabular}
}
\end{table}

\subsection{Performance Comparison with Deep Learning Methods}
To validate the superiority of SES-Adapter, we conducted comparisons with three of the most distinguished sequence-structure hybrid models—MIF-ST \cite{yang2023mif-st}, ESM-GearNet \cite{zhang2023esm-gearnet}, and SaProt-GearNet \cite{su2023saprot}—across seven datasets in Thermostability, MetalIonBinding, GO, and DeepLoc. The baseline scores were derived from the structure-aware language model SaProt \cite{su2023saprot}. The performance score for PLM with SES-Adapter is selected based on the best scores obtained under six different settings and nine baseline models for each downstream task. As shown in Table \ref{tab:hybrid_model}, the PLM equipped with SES-Adapter outperforms these three more complex SOTA hybrid models on most tasks without the need for extensive hyperparameter search. Notably, improvements were significant in MetalIonBinding, BP, and CC, with increases of $3\%$, $11\%$, and $11\%$ respectively.

Additionally, we compared the performance of the SES-Adapter, leveraging \textbf{ESM2-650M}, against non-pretrained methods across three membrane-related datasets. The baseline methods included Moran \cite{feng2000Moran}, DDE \cite{saravanan2015DDE}, ResNet \cite{rao2019tape}, Transformer \cite{rao2019tape}, CNN \cite{shanehsazzadeh2020CNN}, and LSTM \cite{rao2019tape}, with scores derived from PEER's experimental results \cite{xu2022peer}. As depicted in Table \ref{tab:non_pretrain}, integrating pre-trained models with the SES-Adapter significantly improved performance compared to models trained from scratch on downstream tasks (non-pretrained methods). Notably, there was an increase of $5.8\%$ on DeepLocBinary, $1.9\%$ on DeepLocMulti, and $5.3\%$ on DeepSol.

\begin{table}[t]
\caption{Experimental comparison with hybrid model.}
\label{tab:hybrid_model}
\resizebox{\textwidth}{!}{
\begin{tabular}{@{}cccccccc@{}}
\toprule
\multirow{3}{*}{Hybrid Model} & \multirow{2}{*}{Thermostability} & \multirow{2}{*}{MetalIonBinding} & \multicolumn{3}{c}{GO} & \multicolumn{2}{c}{DeepLoc} \\ \cmidrule(l){4-8} 
 &  &  & MF & BP & CC & Multi & Binary \\ \cmidrule(l){2-8} 
 & Spearman’s $\rho$ & ACC ($\%$) & Fmax & Fmax & Fmax & ACC ($\%$) & ACC ($\%$) \\ \midrule
MIF-ST & 0.694 & 75.54 & 0.627 & 0.239 & 0.248 & 78.96 & 91.76 \\
ESM-GearNet & 0.651 & 74.11 & 0.67 & 0.372 & 0.424 & 82.3 & 92.94 \\
SaProt-GearNet & 0.66 & 74.44 & \textbf{0.672}& 0.381 & 0.435 & 84.16 & 93.63 \\ \midrule
SES-Adapter & \textbf{0.704} & \textbf{78.35} & \multicolumn{1}{c}{0.662} & \multicolumn{1}{l}{\textbf{0.489}} & \textbf{0.548} & \textbf{84.54} & \textbf{93.92} \\ \bottomrule
\end{tabular}
}
\end{table}

\begin{table}[t]
\caption{Experimental comparison with non-pre-training method.}
\label{tab:non_pretrain}
\resizebox{0.6\textwidth}{!}{
\begin{tabular}{@{}cccc@{}}
\toprule
Model       & DeepLocBinary  & DeepLocMulti   & DeepSol        \\ \midrule
Moran       & 55.63          & 33.13          & 57.73          \\
DDE         & 77.43          & 49.17          & 59.77          \\
ResNet      & 78.99          & 52.3           & 67.33          \\
Transformer & 75.74          & 56.02          & 70.12          \\
CNN         & 82.67          & 82.67          & 64.43          \\
LSTM        & 88.11          & 62.98          & 70.18          \\ \midrule
SES-Adapter & \textbf{93.92} & \textbf{84.54} & \textbf{75.46} \\ \bottomrule
\end{tabular}
}
\end{table}

\subsection{Training and Convergence Efficiency Analysis}
\begin{table}[t]
\caption{Train loss comparison with vanilla PLM on half-train steps of \textbf{ESM2-150M}. We report \textit{mean (std)} for each dataset.}
\label{tab:speed}
\begin{tabular}{@{}lccc@{}}
\toprule
\multicolumn{1}{c}{Dataset} & Vanilla PLM & SES-Adapter & Speed Difference ($\%$) \\ 
\midrule
DeepLocBinary & 0.2491 & $0.0680_{(0.0053)}$ & 366.4 \\
DeepLocMulti & 0.6972 & $0.1842_{(0.0479)}$ & 378.5 \\
DeepSol & 0.5708 & $0.3430_{(0.0239)}$ & 166.4 \\
DeepSoluE & 0.5740 & $0.4582_{(0.0088)}$ & 125.3 \\
MetalIonBinding & 0.5629 & $0.2322_{(0.0209)}$ & 242.4 \\
Thermotsatbility & 0.0251 & $0.0127_{(0.0007)}$ & 197.9 \\
GO-MF & 0.0231 & $0.0022_{(0.0005)}$ & 1033.7 \\
GO-BP & 0.0279 & $0.0090_{(0.0009)}$ & 308.4 \\
GO-CC & 0.0429 & $0.0099_{(0.0019)}$ & 434.9 \\ 
\bottomrule
\end{tabular}
\end{table}

\begin{figure}[!t]
    \centering
    \includegraphics[width=\textwidth]{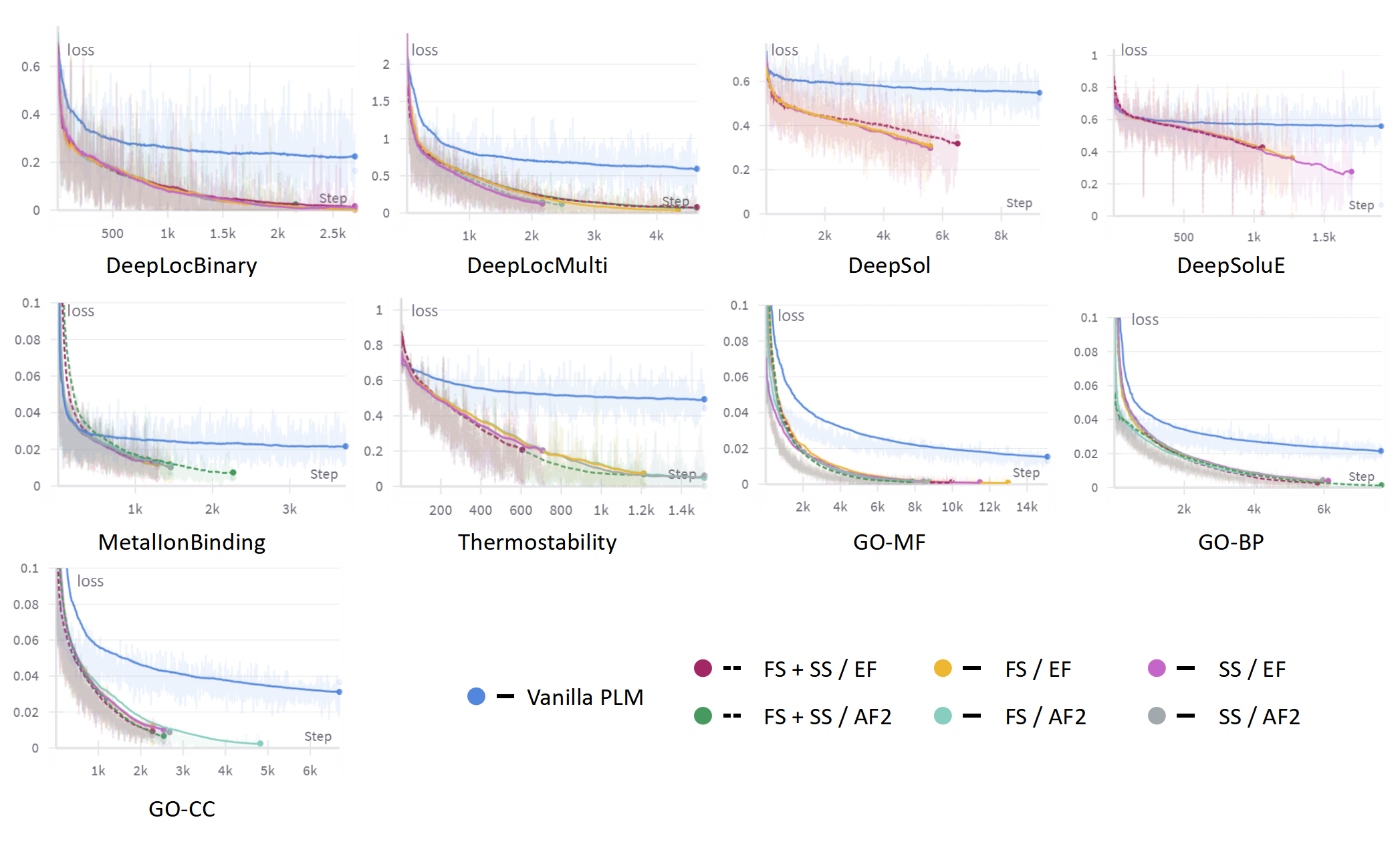}
    \caption{Training loss curve on downstream tasks of \textbf{ESM2-150M}.}
    \label{fig:train_loss}
\end{figure}

\begin{figure}[!t]
    \centering
    \includegraphics[width=\textwidth]{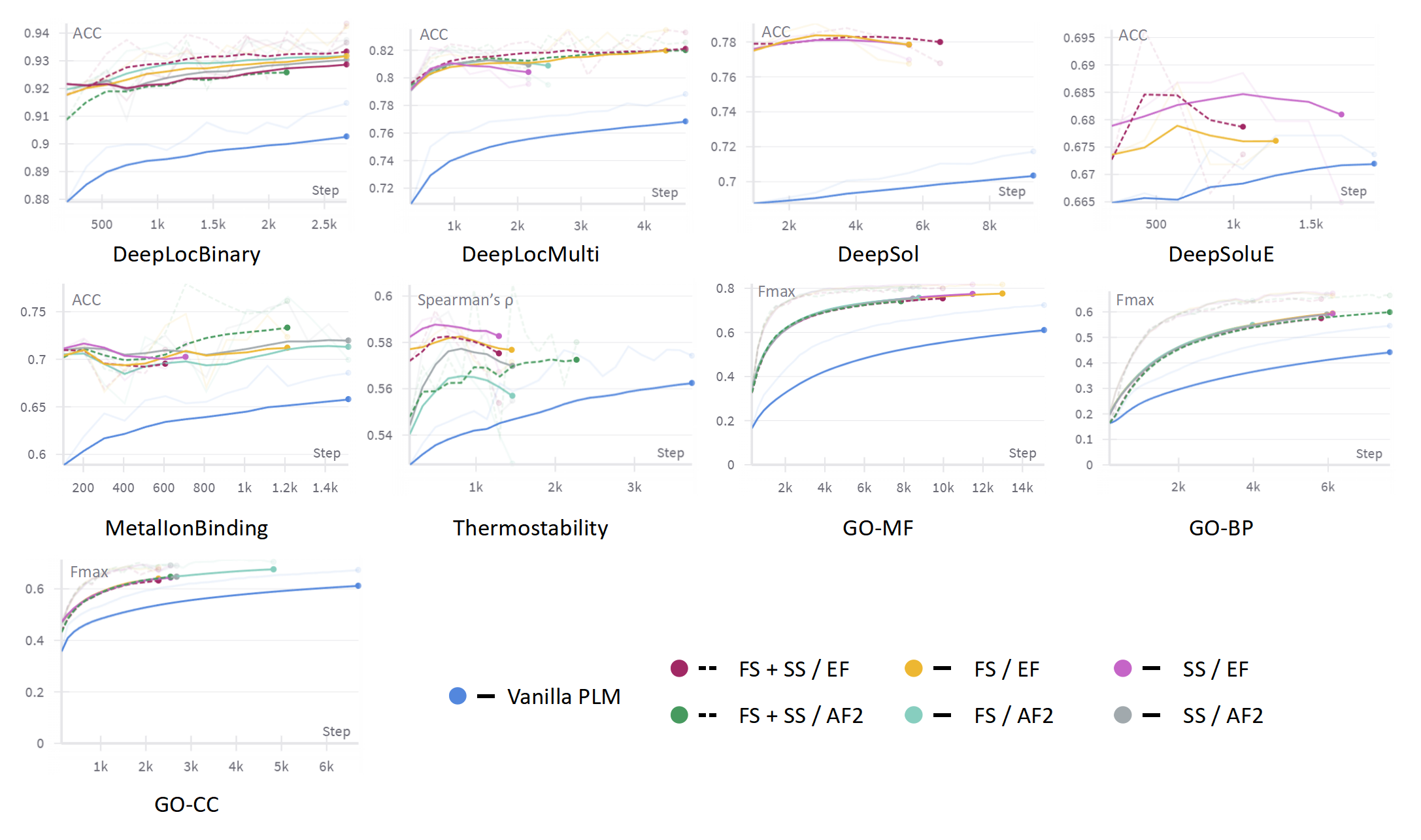}
    \caption{Valid metric curve on downstream tasks of \textbf{ESM2-150M}.}
    \label{fig:valid_metrics}
\end{figure}

To demonstrate the efficiency and rapid convergence of the SES-Adapter, we used \textbf{ESM2-150M} as an example to analyze the training progress. Figure \ref{fig:train_loss} and Figure \ref{fig:valid_metrics} illustrate the differences in training loss reduction and changes in validation set metrics between sessions using the SES-Adapter and those not using it. ``Vanilla" refers to sequence fine-tuning only using PLMs. ``FS" stands for the Foldseek sequence, ``SS" for the DSSP sequence, ``AF2" denotes the SES-Adapter with AlphaFold2 predicted structures, and ``EF" signifies the SES-Adapter with ESMFold predicted structures. Figure \ref{fig:train_loss} shows that the SES-Adapter enhances training efficiency across all downstream tasks, rapidly reducing training loss to very low levels compared to vanilla PLMs. For a more detailed assessment of training efficiency, using half of the training steps of vanilla PLM as a baseline, Table \ref{tab:speed} reveals that training speed increased by up to $1033.7\%$, with the lowest at $125.3\%$, and an average improvement of $361.5\%$ with the use of SES-Adapter. Figure \ref{fig:train_loss} also shows that some configurations using SES-Adapter achieved early convergence, as we adopted validation metrics as a monitor for early stopping, resulting in an approximate 2 times increase in convergence efficiency. Furthermore, Figure \ref{fig:valid_metrics} clearly demonstrates that validation metrics for the SES-Adapter method are consistently and significantly higher than those for vanilla PLMs across all downstream tasks. It is noteworthy that all these curves were plotted and analyzed under the same experimental conditions.

Since the SES-Adapter introduces additional parameters compared to the vanilla PLM, merely comparing the loss at the same number of training steps may not fairly reflect the time cost. Therefore, we conducted repeated runs of the DeepLocBinary and DeepLocMulti datasets on a server without any additional jobs to calculate the time expense per step. On DeepLocBinary, the average was $2.42$ seconds per iteration for the vanilla PLM and $2.53$ seconds per iteration for the SES-Adapter; on DeepLocMulti, the averages were $2.49$ seconds per iteration for the vanilla PLM and $2.64$ seconds per iteration for the SES-Adapter. The training time expenses increased by $4.5\%$ and $6\%$ respectively for these datasets, which is relatively minor and tolerable compared to the training efficiency improvements detailed in Table \ref{tab:speed}.

\subsection{Ablation Study on Components of SES-Adapter}
To validate the contribution of each component designed within the SES-Adapter, we conducted five sets of ablation experiments on three datasets: DeepLocBinary, DeepLocMulti, and DeepSol. The variations included removing the FoldSeek sequence, omitting the DSSP sequence, excluding the RoPE, removing both FoldSeek and DSSP sequences, and utilizing all three components. As shown in Table \ref{tab:ablation}, each component positively contributes to the downstream tasks across the displayed datasets. Performance on downstream tasks was nearly identical when only one type of structural sequence was used. However, omitting structural sequences altogether led to a significant decline in performance, with a nearly 10\% drop on the DeepSol dataset and a 2\% decrease on other datasets, highlighting the SES-Adapter’s ability to efficiently and seamlessly integrate structural information. Compared to the contribution of structural sequences, the contribution from RoPE was relatively minor, fluctuating around 1\%, but it is still evident that position encoding is essential. A more detailed results can be found in Table \textcolor{blue}{S2-S10}.

\begin{table}[t]
\caption{Ablation study on \textbf{ESM2-650M} with different SES-Adapter settings.}
\label{tab:ablation}
\centering
\begin{tabular}{@{}cccccccc@{}}
\toprule
\multirow{2}{*}{FoldSeek} & \multirow{2}{*}{DSSP} & \multirow{2}{*}{RoPE} & \multicolumn{2}{c}{DeepLocBinary} & \multicolumn{2}{c}{DeepLocMulti} & DeepSol \\ \cmidrule(l){4-8} 
 &  &  & AF2 & EF & AF2 & EF & EF \\ \midrule
\ding{55} & \checkmark & \checkmark & 93.00 & 93.35 & 82.09 & 82.24 & 74.21 \\
\checkmark & \ding{55} & \checkmark & 93.87 & 93.32 & 82.42 & 82.31 & 74.13 \\
\checkmark & \checkmark & \ding{55} & 92.87 & 92.52 & 82.46 & 82.02 & 73.61 \\
\ding{55} & \ding{55} & \checkmark & 91.84 & 91.84 & 81.31 & 81.31 & 63.17 \\
\checkmark & \checkmark & \checkmark & \textbf{93.9} & \textbf{93.75} & \textbf{83.41} & \textbf{83.03} & \textbf{74.76} \\ \bottomrule
\end{tabular}%
\end{table}

\subsection{AlphaFold2 and ESMFold Structure Robustness Testing}
\begin{figure}[!t]
    \centering
    \includegraphics[width=\textwidth]{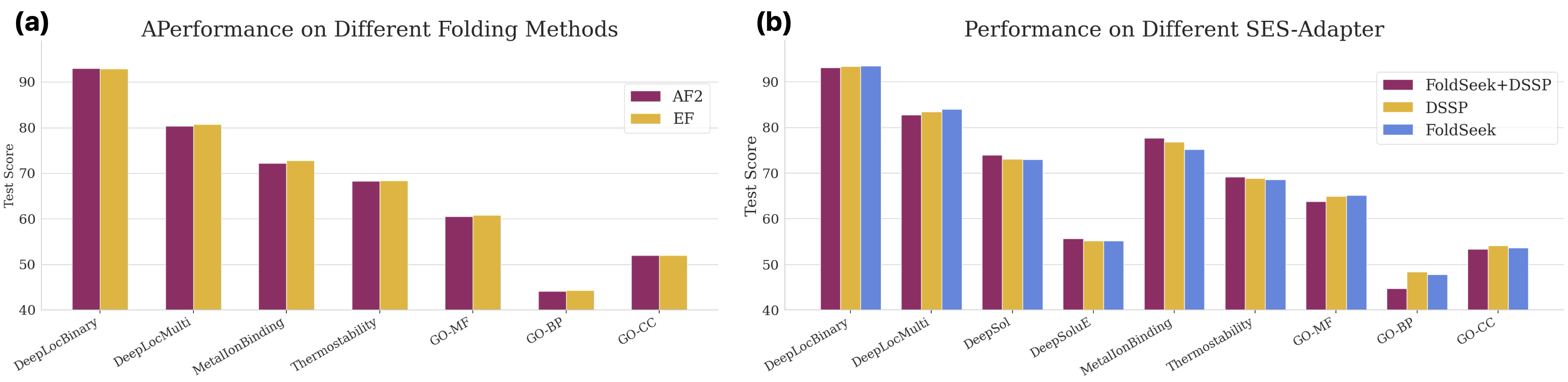}
    \caption{(a) Average performance of AlphaFold2 and ESMFold on downstream tasks. (b) Average performance of different SES-Adapter settings on downstream tasks.}
    \label{fig:ablation}
\end{figure}

Different protein folding methods yield structures of varying quality. We aimed to determine whether using serialized structures could mitigate potential errors caused by low-quality structures. According to Table \ref{tab:dataset}, the pLDDT score differences between two structures are minimal for the DeepLocBinary, DeepLocMulti, and Thermostability datasets, whereas the MetalIonBinding, GO-MF, GO-BP, and GO-CC datasets show pLDDT score differences greater than $10$ between the two structures. Figure \ref{fig:ablation} (a) presents the average scores across nine models and three SES-Adapter settings, using both structures across various datasets. It is observed that the quality of the structure has virtually no impact on the performance of downstream tasks, with the maximum difference being $0.6\%$ and the minimum difference being $0.07\%$. This confirms that the SES-Adapter has excellent robustness to the protein structure quality.

\subsection{Importance of FoldSeek and DSSP Structural Sequences}

The interaction attention between structural sequences and amino acid sequences is a highlight of the SES-Adapter, and it is crucial to determine which types of structural information most significantly enhance the quality of PLM embeddings and assist in improving downstream tasks. Figure \ref{fig:ablation} (b) shows the average scores across nine models and two structures under various SES-Adapter configurations, where DeepSol and DeepSoluE represent the average scores for the nine models using ESMFold structures. From Figure \ref{fig:ablation} (b), it can be observed that the optimization varies for different tasks. For example, using both FoldSeek and DSSP sequences together yields the best results in the MetalIonBinding task, but performs moderately in the three branches of the GO dataset. Overall, using both types of information achieved the best results in four datasets, using just the FoldSeek sequence was optimal in three datasets, and using the DSSP sequence was best in two datasets. Therefore, incorporating diverse structural information generally provides a beneficial gain for downstream tasks.

\section{Conclusion and Discussion}

Using large model representations to aid predictions in downstream tasks is a common and effective approach. Inspired by the adapter methods used in the NLP field, this study introduces a simple, efficient, and scalable adapter architecture for PLMs, which minimally introduces additional parameters. By incorporating structural sequences for cross-modal fusion attention, this adapter further enhances the representational quality of PLMs to improve downstream tasks. This architecture is not limited to using FoldSeek and DSSP; it can also adopt other more effective structural serialization methods. The design is straightforward, requires no manual feature engineering, facilitates easy parameter tuning, reduces the need for prior knowledge, and can be easily extended to different model architectures, significantly enhancing training and convergence efficiency while overcoming the negative optimization issues caused by errors in predicted structures. We conducted extensive testing and validation across nine models, nine datasets, and two types of folding structures, achieving favorable results in each task.

However, the SES-Adapter is still relatively nascent in the domain of fine-tuning large protein models, demonstrating the feasibility and superiority of structure adapters in the PLM field. Future work could explore more sophisticated methods for integrating sequences and structures in fine-tuning, or develop more complex and effective methods for serializing protein structures, such as incorporating the topological spatial structures of each amino acid into the serialization process. Indeed, while structure dictates function, the essence remains the sequence of information. Future explorations could also focus on enhancing PLM representational capabilities and downstream task performance through protein vector retrieval methods.

\begin{acknowledgement}

This work was supported by Research Programme of National Engineering Laboratory for Big Data Distribution and Exchange Technologies, Shanghai Municipal Special Fund for Promoting High Quality Development (No. 2021-GYHLW-01007), the National Natural Science Foundation of China (11974239, 12104295), the Innovation Program of Shanghai Municipal Education Commission (2019-01-07-00-02-E00076), Shanghai Jiao Tong University Scientific and Technological Innovation Funds (21X010200843), the Student Innovation Center at Shanghai Jiao Tong University, and Shanghai Artificial Intelligence Laboratory.

\end{acknowledgement}

\begin{suppinfo}

The data and code are available at \url{https://github.com/tyang816/SES-Adapter}. Table \textcolor{blue}{S1}. The detailed implementation information about baseline methods. Table \textcolor{blue}{S2-S10}. Detailed results of different task datasets.

\end{suppinfo}

\bibliography{JCIM/ACS_main}

\end{document}








\maketitle

\listoftables

\newpage

\begin{table}[]
\centering
\resizebox{\textwidth}{!}{%
\begin{tabular}{@{}cccl@{}}
\toprule
Model & Version & \# Params (Millions) & \multicolumn{1}{c}{Implementation} \\ \midrule
\multicolumn{4}{c}{\textbf{Encoder-Only}} \\ \midrule
 & t30 & 150 & \url{https://huggingface.co/facebook/esm2_t30_150M_UR50D} \\
 & t33 & 652 & \url{https://huggingface.co/facebook/esm2_t33_650M_UR50D} \\
\multirow{-3}{*}{ESM2} & t36 & 3,000 & \url{https://huggingface.co/facebook/esm2_t36_3B_UR50D} \\ \midrule
ESM-1b & - & 652 & \url{https://huggingface.co/facebook/esm1b_t33_650M_UR50S} \\ \midrule
 & uniref & 420 & \url{https://huggingface.co/Rostlab/prot_bert} \\
\multirow{-2}{*}{ProtBert} & bfd & 420 & \url{https://huggingface.co/Rostlab/prot_bert_bfd} \\ \midrule
\multicolumn{4}{c}{\textbf{Encoder-Decoder}} \\ \midrule
 & xl\_uniref & 3,000 & \url{https://huggingface.co/Rostlab/prot_t5_xl_uniref50} \\
\multirow{-2}{*}{ProtT5} & xl\_bfd & 3,000 & \url{https://huggingface.co/Rostlab/prot_t5_xl_bfd} \\ \midrule
 & base & 450M & \url{https://huggingface.co/ElnaggarLab/ankh-base} \\
\multirow{-2}{*}{Ankh} & large & 1,150 & \url{https://huggingface.co/ElnaggarLab/ankh-large} \\ \bottomrule
\end{tabular}%
}
\caption{Summary of baseline methods.}
\label{tab:my-table}
\end{table}

\newpage

\begin{table}[]
\centering
\resizebox{0.9\textwidth}{!}{%
\begin{tabular}{@{}ccccccccc@{}}
\toprule
\multicolumn{2}{c}{\multirow{2}{*}{DeepLocBinary}} & \multicolumn{3}{c}{AlphaFold2} & \multicolumn{3}{c}{EsmFold} & None \\ \cmidrule(l){3-9} 
\multicolumn{2}{c}{} & FS+SS & SS & FS & FS+SS & SS & FS & - \\ \midrule
\multirow{3}{*}{ESM2} & t30\_UR50D & 93.34 & 93.23 & 92.59 & 93.75 & 93.4 & 93.06 & 91.2 \\
 & t33\_UR50D & 93.9 & 93 & 93.87 & 93.75 & 93.35 & 93.32 & 91.84 \\
 & t36\_UR50D & 93.11 & 93.46 & 93.23 & 93.52 & 93.4 & 93.23 & 90.57 \\ \midrule
\multirow{2}{*}{ProtBert} & prot\_bert & 92.59 & 91.09 & 91.61 & 91.26 & 91.38 & 91.09 & 87.09 \\
 & prot\_bert\_bfd & 91.86 & 93.11 & 92.01 & 92.07 & 92.19 & 91.44 & 89.01 \\ \midrule
\multirow{2}{*}{ProtT5} & xl\_uniref50 & 93.29 & 93.58 & 93.29 & 92.82 & 93.34 & 93.52 & 92.25 \\
 & xl\_bfd & 93.11 & 93.23 & 92.71 & 92.77 & 91.72 & 92.77 & 91.72 \\ \midrule
\multirow{2}{*}{Ankh} & base & 93.63 & 92.77 & 92.82 & 93.75 & 93.4 & 93.58 & 89.76 \\
 & large & 93.52 & 93.34 & 92.94 & 92.71 & 93.46 & 93.92 & 89.93 \\ \bottomrule
\end{tabular}%
}
\caption{Detailed results of DeepLocBinary.}
\label{tab:DeepLocBinary}
\end{table}

\newpage

\begin{table}[]
\centering
\resizebox{0.9\textwidth}{!}{%
\begin{tabular}{@{}ccccccccc@{}}
\toprule
\multicolumn{2}{c}{\multirow{2}{*}{DeepLocMulti}} & \multicolumn{3}{c}{AlphaFold2} & \multicolumn{3}{c}{EsmFold} & None \\ \cmidrule(l){3-9} 
\multicolumn{2}{c}{} & FS+SS & SS & FS & FS+SS & SS & FS & - \\ \midrule
\multirow{3}{*}{ESM2} & t30\_UR50D & 78.48 & 80.78 & 80.01 & 78.81 & 80.53 & 78.52 & 77.02 \\
 & t33\_UR50D & 83.41 & 82.09 & 82.42 & 83.03 & 82.24 & 82.31 & 81.31 \\
 & t36\_UR50D & 81.04 & 81.62 & 83.19 & 83.48 & 83.66 & 82.42 & 80.82 \\ \midrule
\multirow{2}{*}{ProtBert} & prot\_bert & 74.51 & 75.57 & 73.71 & 76.19 & 75.82 & 75.24 & 74.14 \\
 & prot\_bert\_bfd & 76.73 & 77.86 & 76.51 & 77.28 & 77.57 & 77.43 & 75.2 \\ \midrule
\multirow{2}{*}{ProtT5} & xl\_uniref50 & 83.92 & 83.3 & 82.97 & 83.88 & 83.33 & 84.54 & 82.02 \\
 & xl\_bfd & 78.91 & 79.65 & 80.56 & 79.97 & 79.8 & 80.67 & 78.88 \\ \midrule
\multirow{2}{*}{Ankh} & base & 81.44 & 81.25 & 80.27 & 77.97 & 81.98 & 83.15 & 78.48 \\
 & large & 82.31 & 83.59 & 84.43 & 83.22 & 83.3 & 83.55 & 78.81 \\ \bottomrule
\end{tabular}%
}
\caption{Detailed results of DeepLocMulti.}
\label{tab:DeepLocMulti}
\end{table}

\newpage

\begin{table}[]
\centering
\resizebox{0.9\textwidth}{!}{%
\begin{tabular}{@{}ccccccccc@{}}
\toprule
\multicolumn{2}{c}{\multirow{2}{*}{MetalIonBinding}} & \multicolumn{3}{c}{AlphaFold} & \multicolumn{3}{c}{EsmFold} & None \\ \cmidrule(l){3-9} 
\multicolumn{2}{c}{} & FS+SS & SS & FS & FS+SS & SS & FS & - \\ \midrule
\multirow{3}{*}{ESM2} & t30\_UR50D & 72.03 & 71.73 & 71.28 & 73.83 & 72.48 & 72.03 & 68.57 \\
 & t33\_UR50D & 71.58 & 73.08 & 71.43 & 72.78 & 69.92 & 69.17 & 67.97 \\
 & t36\_UR50D & 71.28 & 71.58 & 69.77 & 70.98 & 74.74 & 71.28 & 70.24 \\ \midrule
\multirow{2}{*}{ProtBert} & prot\_bert & 68.87 & 67.52 & 68.12 & 72.48 & 67.22 & 66.47 & 64.96 \\
 & prot\_bert\_bfd & 69.32 & 66.62 & 73.08 & 69.47 & 68.27 & 71.73 & 65.41 \\ \midrule
\multirow{2}{*}{ProtT5} & xl\_uniref50 & 77.29 & 74.29 & 75.64 & 76.69 & 78.2 & 73.68 & 75.24 \\
 & xl\_bfd & 72.78 & 75.04 & 71.73 & 72.93 & 76.39 & 74.29 & 73.14 \\ \midrule
\multirow{2}{*}{Ankh} & base & 74.74 & 71.02 & 72.78 & 73.53 & 72.83 & 73.08 & 72.64 \\
 & large & 77.89 & 75.34 & 74.29 & 77.44 & 78.35 & 76.09 & 73.83 \\ \bottomrule
\end{tabular}%
}
\caption{Detailed results of MetalIonBinding.}
\label{tab:MetalIonBinding}
\end{table}

\newpage

\begin{table}[]
\centering
\resizebox{0.9\textwidth}{!}{%
\begin{tabular}{@{}ccccccccc@{}}
\toprule
\multicolumn{2}{c}{\multirow{2}{*}{Thermostability}} & \multicolumn{3}{c}{AlphaFold} & \multicolumn{3}{c}{EsmFold} & None \\ \cmidrule(l){3-9} 
\multicolumn{2}{c}{} & FS+SS & SS & FS & FS+SS & SS & FS & - \\ \midrule
\multirow{3}{*}{ESM2} & t30\_UR50D & 68.99 & 69.78 & 68.93 & 70.21 & 68.53 & 68.99 & 68.69 \\
 & t33\_UR50D & 68.78 & 69.28 & 68.82 & 69.85 & 70.44 & 69.66 & 67.31 \\
 & t36\_UR50D & 68.86 & 69.4 & 69.42 & 68.96 & 68.94 & 68.18 & 68.53 \\ \midrule
\multirow{2}{*}{ProtBert} & prot\_bert & 65.84 & 65.5 & 65.98 & 65.62 & 65.82 & 64.85 & 65.35 \\
 & prot\_bert\_bfd & 66.89 & 66.95 & 66.42 & 65.38 & 65.66 & 65.54 & 65.28 \\ \midrule
\multirow{2}{*}{ProtT5} & xl\_uniref50 & 69.68 & 68.93 & 68.89 & 69.48 & 68.87 & 68.63 & 68.36 \\
 & xl\_bfd & 66.82 & 67.79 & 67.35 & 68.87 & 68.68 & 68.41 & 67.29 \\ \midrule
\multirow{2}{*}{Ankh} & base & 68.84 & 69.04 & 68.67 & 70.37 & 70.11 & 69.81 & 68.33 \\
 & large & 69.55 & 69.02 & 68.58 & 68.74 & 68.73 & 68.57 & 67.52 \\ \bottomrule
\end{tabular}%
}
\caption{Detailed results of Thermostability.}
\label{tab:Thermostability}
\end{table}

\newpage

\begin{table}[]
\centering
\resizebox{0.9\textwidth}{!}{%
\begin{tabular}{@{}ccccccccc@{}}
\toprule
\multicolumn{2}{c}{\multirow{2}{*}{MF}} & \multicolumn{3}{c}{AlphaFold} & \multicolumn{3}{c}{EsmFold} & None \\ \cmidrule(l){3-9} 
\multicolumn{2}{c}{} & FS+SS & SS & FS & FS+SS & SS & FS & - \\ \midrule
\multirow{2}{*}{ESM2} & t30\_UR50D & 61.06 & 61.29 & 61.8 & 61.79 & 62.01 & 62.14 & 56.44 \\
 & t33\_UR50D & 62.54 & 62.32 & 61.79 & 62.49 & 62.12 & 62.66 & 60.8 \\ \midrule
\multirow{2}{*}{ProtBert} & prot\_bert & 51.98 & 52.61 & 53.4 & 52.25 & 52.95 & 53.55 & 46.75 \\
 & prot\_bert\_bfd & 56.08 & 58.62 & 57.71 & 57.27 & 58.68 & 57.64 & 46.59 \\ \midrule
\multirow{2}{*}{Ankh} & base & 63.07 & 65.63 & 66.23 & 63.51 & 66 & 65.92 & 57.34 \\
 & large & 63.51 & 64.88 & 65.09 & 63.98 & 64.95 & 65.23 & 55.64 \\ \bottomrule
\end{tabular}%
}
\caption{Detailed results of GO-MF.}
\label{tab:GO-MF}
\end{table}

\newpage

\begin{table}[]
\centering
\resizebox{0.9\textwidth}{!}{%
\begin{tabular}{@{}ccccccccc@{}}
\toprule
\multicolumn{2}{c}{\multirow{2}{*}{BP}} & \multicolumn{3}{c}{AlphaFold} & \multicolumn{3}{c}{EsmFold} & None \\ \cmidrule(l){3-9} 
\multicolumn{2}{c}{} & FS+SS & SS & FS & FS+SS & SS & FS & - \\ \midrule
\multirow{2}{*}{ESM2} & t30\_UR50D & 43.11 & 44.83 & 43.88 & 43.21 & 45.44 & 45.04 & 41.21 \\
 & t33\_UR50D & 46.31 & 45.56 & 46.12 & 47.04 & 46.54 & 45.97 & 45.46 \\ \midrule
\multirow{2}{*}{ProtBert} & prot\_bert & 39.28 & 39.32 & 38.94 & 39.05 & 39.09 & 39.27 & 38.24 \\
 & prot\_bert\_bfd & 39.59 & 42.04 & 43.11 & 39.01 & 41.07 & 43.16 & 38.94 \\ \midrule
\multirow{2}{*}{Ankh} & base & 45.14 & 48.86 & 48.16 & 45.68 & 48.88 & 48.11 & 42.23 \\
 & large & 44.65 & 48.39 & 47.66 & 44.83 & 48.3 & 47.91 & 41.72 \\ \bottomrule
\end{tabular}%
}
\caption{Detailed results of GO-BP.}
\label{tab:GO-BP}
\end{table}

\newpage

\begin{table}[]
\centering
\resizebox{0.9\textwidth}{!}{%
\begin{tabular}{@{}ccccccccc@{}}
\toprule
\multicolumn{2}{c}{\multirow{2}{*}{CC}} & \multicolumn{3}{c}{AlphaFold} & \multicolumn{3}{c}{EsmFold} & None \\ \cmidrule(l){3-9} 
\multicolumn{2}{c}{} & FS+SS & SS & FS & FS+SS & SS & FS & - \\ \midrule
\multirow{2}{*}{ESM2} & t30\_UR50D & 50.58 & 50.18 & 51.26 & 50.54 & 50.96 & 51.73 & 48.11 \\
 & t33\_UR50D & 52.58 & 54.79 & 51.51 & 50.87 & 52.93 & 53.8 & 51.23 \\ \midrule
\multirow{2}{*}{ProtBert} & prot\_bert & 50.36 & 52.13 & 50.96 & 50.82 & 53.05 & 50.53 & 49.89 \\
 & prot\_bert\_bfd & 50.29 & 51.41 & 51.33 & 50.55 & 52.12 & 51.2 & 50.05 \\ \midrule
\multirow{2}{*}{Ankh} & base & 51.56 & 53.15 & 52.81 & 50.83 & 51.85 & 53.02 & 50.85 \\
 & large & 53.32 & 53.75 & 54.54 & 53.29 & 54.52 & 52.66 & 52.77 \\ \bottomrule
\end{tabular}%
}
\caption{Detailed results of GO-CC.}
\label{tab:GO-CC}
\end{table}

\newpage

\begin{table}[]
\centering
\resizebox{0.65\textwidth}{!}{%
\begin{tabular}{@{}cccccc@{}}
\toprule
\multicolumn{2}{c}{\multirow{2}{*}{DeepSol}} & \multicolumn{3}{c}{EsmFold} & None \\ \cmidrule(l){3-6} 
\multicolumn{2}{c}{} & FS+SS & SS & FS & - \\ \midrule
\multirow{3}{*}{ESM2} & t30\_UR50D & 74.46 & 74.61 & 73.81 & 63.37 \\
 & t33\_UR50D & 74.76 & 74.21 & 74.13 & 63.17 \\
 & t36\_UR50D & 73.96 & 72.76 & 72.61 & 64.17 \\ \midrule
\multirow{2}{*}{ProtBert} & prot\_bert & 72.96 & 72.36 & 71.51 & 63.02 \\
 & prot\_bert\_bfd & 74.46 & 73.91 & 72.71 & 64.32 \\ \midrule
\multirow{2}{*}{ProtT5} & xl\_uniref50 & 75.46 & 73.06 & 73.01 & 67.07 \\
 & xl\_bfd & 74.21 & 71.66 & 73.06 & 66.17 \\ \midrule
\multirow{2}{*}{Ankh} & base & 72.86 & 72.41 & 72.31 & 62.57 \\
 & large & 72.01 & 73.11 & 73.41 & 63.72 \\ \bottomrule
\end{tabular}%
}
\caption{Detailed results of DeepSol.}
\label{tab:DeepSol}
\end{table}

\newpage

\begin{table}[]
\centering
\resizebox{0.65\textwidth}{!}{%
\begin{tabular}{@{}cccccc@{}}
\toprule
\multicolumn{2}{c}{\multirow{2}{*}{DeepSoluE}} & \multicolumn{3}{c}{EsmFold} & None \\ \cmidrule(l){3-6} 
\multicolumn{2}{c}{} & FS+SS & SS & FS & - \\ \midrule
\multirow{3}{*}{ESM2} & t30\_UR50D & 56.12 & 55.03 & 54.45 & 55.12 \\
 & t33\_UR50D & 55.52 & 55.97 & 56.97 & 55.97 \\
 & t36\_UR50D & 54.94 & 54.52 & 54.74 & 54.58 \\ \midrule
\multirow{2}{*}{ProtBert} & prot\_bert & 55.29 & 54.23 & 54.42 & 54.24 \\
 & prot\_bert\_bfd & 56.1 & 56.06 & 55.55 & 54.97 \\ \midrule
\multirow{2}{*}{ProtT5} & xl\_uniref50 & 55.74 & 55.23 & 55.32 & 55.03 \\
 & xl\_bfd & 55.13 & 55.87 & 55.71 & 55.32 \\ \midrule
\multirow{2}{*}{Ankh} & base & 55.97 & 54.94 & 55.16 & 55.15 \\
 & large & 55.87 & 54.55 & 54.39 & 54.25 \\ \bottomrule
\end{tabular}%
}
\caption{Detailed results of DeepSoluE.}
\label{tab:DeepSoluE}
\end{table}